\documentclass{article}

\usepackage{arxiv}

\usepackage{array}
\usepackage[flushleft]{threeparttable}
\usepackage[table]{xcolor}
\usepackage{cite}
\def\BibTeX{{\rm B\kern-.05em{\sc i\kern-.025em b}\kern-.08em
    T\kern-.1667em\lower.7ex\hbox{E}\kern-.125emX}}

\usepackage[utf8]{inputenc} 
\usepackage[T1]{fontenc}    
\usepackage{hyperref}       
\hypersetup{
	colorlinks=true,
	linkcolor=blue,
	filecolor=magenta,      
	urlcolor=cyan,
}
\usepackage{url}            
\usepackage{booktabs}       
\usepackage{amsfonts}       
\usepackage{nicefrac}       
\usepackage{microtype}      
\usepackage{subcaption}
\usepackage{graphicx}
\usepackage{textcomp, setspace}
\usepackage{amsmath}
\usepackage{authblk}
\title{
	A Novel Transformer Network with Shifted Window Cross-Attention for Spatiotemporal Weather Forecasting
}
\author{Alabi Bojesomo}
\author{Hasan Al Marzouqi}
\author{Panos Liatsis}

\affil{Khalifa University, Abu Dhabi, UAE \authorcr
	\{\tt alabi.bojesomo, hasan.almarzouqi, panos.liatsis\}@ku.ac.ae}

\begin{document}
\maketitle

\begin{abstract}
Earth Observatory is a growing research area that can capitalize on the powers of AI for short time forecasting, a Now-casting scenario. In this work, we tackle the challenge of weather forecasting using a video transformer network. Vision transformer architectures have been explored in various applications, with major constraints being the computational complexity of Attention and the data hungry training. To address these issues, we propose the use of Video Swin-Transformer, coupled with a dedicated augmentation scheme. Moreover, we employ gradual spatial reduction on the encoder side and cross-attention on the decoder. The proposed approach is tested on the Weather4Cast2021 weather forecasting challenge data, which requires the prediction of 8 hours ahead future frames (4 per hour) from an hourly weather product sequence.
The dataset was normalized to 0-1 to facilitate using the evaluation metrics across different datasets.
The model results in an MSE score of 0.4750 when provided with training data, and 0.4420 during transfer learning without using training data, respectively.
\end{abstract}


\section{Introduction}
Forecasting the weather condition is a crucial driving force in agriculture and autonomous vehicles industry \cite{REN2021100178_weather_survey}. Accurate weather forecast affects the successful deployment of autonomous vehicles and management of food production.
When designing autonomous navigation and collision avoidance technologies, for example, awareness of the weather is an essential part of the location context.
Also, adequate weather prediction is essential for monitoring soil nutrients and regulating crop yield in the agricultural industry.

Artificial intelligence (AI) is gradually gaining traction in weather forecasting  due to its relative simplicity, when compared to numerical weather prediction (NWP) \cite{fente2018weather, wang2019deep, singh2019weather,hewage2020temporal,  hewage2021deep}.
AI approaches applicable to weather forecasting can be categorized based on the network structure, e.g., Convolutional Neural Networks (CNN) \cite{qiu2017a, nascimento-a, yonekura2018a, Zhang_deep2021}, Autoencoders \cite{hossain2015a, lin2018a}, and recurrent networks \cite{s-a, karevan-a, Hou_LSTM, Taisong_LSTM}.

CNN has been used in many state-of-the-art image classification \cite{Alexnet, resnet, densenet} and semantic segmentation models \cite{unet, bojesomo2020traffic}.
Spatiotemporal forecasting has equally been tackled with CNN models, based on their demonstrated performance on segmentation tasks \cite{choi2020utilizing, t4c2020-kopp21a, bojesomo2020traffic}. 
Recently, models based on self attention transformer are gradually revolutionizing computer vision applications \cite{Dosovitskiy2021_image_transformer, Wang2021PyramidVT, ranftl2021vision_densePrediction,dai2021coatnet}. 
While Natural language processing (NLP) uses transformer networks on encoded tokens, vision transformer-based models utilize patch-based image encoding \cite{Dosovitskiy2021_image_transformer}. 
Vision transformer is showing promise in many applications, however, it involves computationally costly self attention, which poses a challenge to be addressed.
A number of modifications of self attention \cite{dong2021cswin, Liu2021_swin_transformer, Liu2021_video_swin_transformer, wang2021crossformer} have been proposed in the literature to address the computational demands of self attention.

Vision transformer is becoming popular in computer vision applications, including semantic segmentation among others \cite{transformer_vaswani2017, Dosovitskiy2021_image_transformer, Wang2021PyramidVT, ranftl2021vision_densePrediction}.
\cite{Dosovitskiy2021_image_transformer} uses patch-based image encoding, similar to word embedding in NLP, making vision amenable to transformer network \cite{transformer_vaswani2017}. 
Other works explore modifications of the attention layer, resulting in more efficient transformer architectures for vision tasks, such as the pyramid vision transformer \cite{Wang2021PyramidVT}. 
Spatiotemporal forecasting is similar to dense prediction tasks, e.g., depth estimation and segmentation, which have been tackled by CNN based architectures \cite{unet}. More recently, however, vision transformer is gradually becoming the dominant architectures in these applications \cite{ranftl2021vision_densePrediction, Wang2021PyramidVT, Dosovitskiy2021_image_transformer, Liu2021_swin_transformer}.

Among the widely adopted vision transformer architecture is the \textit{Shifted Window Transformer} (Swin Transformer), which uses local window attention with shifting operation between transformer layers \cite{Liu2021_swin_transformer}, popular in classification tasks. Swin transformer has also been used as encoder (feature extractor) for downstream tasks, such as object detection, segmentation \cite{Cao2021_swin_unet}, and forecasting \cite{bojesomo2020traffic, bojesomo2021spatiotemporal, bojesomo2021swindecoder, bojesomo2022swinunet3d}.
\cite{Cao2021_swin_unet} replace all CNN blocks of the typical UNet architecture \cite{unet} with Swin transformer blocks, resulting in the Swin-UNet architecture. 
Swin-UNet uses an MLP (Multilayer perceptron)-based layer (patch expanding layer) for upsampling in the decoding branch, making it exclude averaging based upsampling used in UNet while being the opposite of the MLP-based downsampling layer (patch merging layer) used in pyramid vision transformers  \cite{Cao2021_swin_unet, Liu2021_swin_transformer, Wang2021PyramidVT}.
\cite{Liu2021_video_swin_transformer} proposed the 3-dimensional (3D/Video) Swin Transformer, replacing patch embedding, patch merging, and shifted window transformer with their respective 3D variants, resulting in a parameter efficient network, as evidenced in performance on several video datasets (e.g., Something-Something v2 \cite{goyal2017something}, Kinetics-400 \cite{kay2017kinetics}, and kinetics-600) \cite{Liu2021_video_swin_transformer}.

In this research, to the best of the authors' knowledge, the video Swin transformer network \cite{Liu2021_video_swin_transformer} is used for the first time in short term weather forecasting. Moreover, the following improvements in the video Swin transformer architecture are proposed:
\begin{itemize}
    \item \textbf{3D patch expanding} is used instead of the widely used upsampling layer in the decoder. Fully connected layers are used to ensure appropriate learning of the upsampled data.
    \item \textbf{Cross attention block} is introduced in the shifted window transformer decoder. This block is introduced to better represent the interrelationships in the upsampled data in between the decoder and the skipped connected features from the encoder branch.
    \item We propose a carefully designed data augmentation scheme, alleviating the need for pre-training of the transformer-based network. 
\end{itemize}

Moreover, we build upon the preliminary results of a previous study \cite{bojesomo2021swindecoder, bojesomo2021spatiotemporal} by using a much larger dataset. We provide an ablation study on both the data and network structure with a quantitative analysis and systematic evaluation of model prediction performance.

The outline of the paper is as follows. A summary of related works is presented in Section \ref{sec: related works}. The details of the proposed architecture and its layers are given in Section \ref{sec: methods}, while the experimental results of the proposed approach in the IEEE Big Data 2021 Weather4cast challenge data are presented in Section \ref{sec: result}. Finally, Section \ref{sec: conclusion} provides the conclusions of this research and avenues for further work.

\section{Related Works}
\label{sec: related works}
Weather forecasting is the foundation of meteorology. Traditionally, weather prediction is based on the physical modelling of the associated physical phenomena. This numerical weather prediction (NWP) approach requires the solution of spatiotemporal partial differential equations (PDE), related to a variety of underlying atmospheric processes, including radiative, chemical, dynamic, thermodynamic, etc \cite{bauer2015a}. It is obvious that modelling and solving of this multitude of atmospheric processes requires substantial computational resources, which is, indeed, one of the disadvantages of the NWP method \cite{sandu2013so}. However, the most important drawback of the PDE approach is the chaotic nature of weather \cite{abraham2002a}, which makes model performance highly dependent on the initial conditions. Whilst the use of NWP in short-term weather forecasting is problematic, it is a popular approach in long-term forecasting, as it is able to track long-scale trends \cite{voyant2012numerical}. 

The spatiotemporal nature of weather prediction offers major opportunities and challenges for the use of AI and data analytics methods. 
Data-driven weather prediction helps to avoid both the known and unknown chaotic behavior of weather fluctuations by focusing on the evolution of the available data \cite{pathak2018model}. 
Physical phenomena associated with the weather are too complex to model and sometimes not well understood, paving the way for a relatively simpler, data-driven approach.
AI methods for weather forecasting can be classified into machine learning- and deep learning-based. 
Furthermore, machine learning (ML) architectures, applied to weather forecasting, can be grouped into static and dynamic (recurrent) in terms of whether the prediction is sequentially generated \cite{He2021, Bochenek2022, Juarez2022, Achite2022}, or not. Static ML architectures include clustering (K-means, PCA) \cite{fang2021classification, Lou2021}, artificial neural networks (ANN) \cite{abbot2017application, Bochenek2022, Almikaeel2022},  graph neural networks (GNN) \cite{ma2022histgnn}, clustering + neural networks \cite{shukla2013prediction, lan2019day}, decision trees such as Gradient Boosting (XGBoost \cite{huang2021solar}, AdaBoost \cite{Juarez2022}, CatBoost \cite{Niu2021, Monego2022, Juarez2022}), and Random Forest \cite{iverson2004new, Yu201792}. Dynamic ML architectures for weather forecasting, on the other hand, make use of the sequential nature of the training data in generating the forecast. Methods in this category include partially recurrent architectures (e.g., Elman neural networks (ENN)) \cite{Song2021}, and fully recurrent ones, such as recurrent neural networks (RNN), long short-term memory (LSTM) and gated recurrent units (GRU). The principle of fully recurrent architectures is that they capitalize on the dynamic patterns in the input data sequences  \cite{liu2014deep, LIN2018446, hossain2015a, karevan-a, nascimento-a, leinonen2021improvements}. The common challenge of machine learning-based methods is that they require a good understanding of weather parameters in order to perform appropriate feature engineering. This limits their applicability as not all contributing factors to weather changes are yet known or can be accurately measured and/or tracked.

Deep learning-based approaches alleviate the need for the feature engineering stage of ML methods. Instead, feature learning is automatically performed with the use of convolution blocks, which allows for data-driven extraction of the required features. These features are then used as the input to the classification/ regression layer, which is usually a fully connected layer. Considering the spatiotemporal setting of weather data, CNN can be readily applied \cite{Zhang_deep2021}. 
In \cite{unet}, a UNet  architecture with densely connected backbone was used for weather prediction \cite{sungbin_weather_stage2_2021}, where the continuous aggregation of the densely connected CNN in the backbone ensures reuse of the intermediate (state) results. Moreover, generative adversarial networks (GAN) have been successfully applied in weather forecasting  \cite{kwok_weather_stage2_2021, ravuri2021skilful, Wang_GAN}. 
Spatiotemporal weather forecasting using deep learning requires appropriate treatment of the spatial and temporal information in the data.
For instance, CNN can be used to extract features from the spatial information, while recurrent networks (RNN, LSTM, GRU) can be used to model the temporal interrelationship.
Combining CNN with recurrent networks results in architectures, such as ConvLSTM \cite{shi2015, Taisong_LSTM, Hou_LSTM}, PredRNN \cite{predrnn}, PredRNN++ \cite{pmlr-v80-wang18b}, MetNet \cite{s-a}, TrajGRU \cite{shi2017}, ConvGRU \cite{leinonen2021improvements}. 
Spatiotemporal weather forecasting can also be tackled with transformers, originally proposed for NLP \cite{transformer_vaswani2017}. This is because, attention networks used in transformer can eradicate the need for time consuming sequential forecasting, an approach common to recurrent networks (ConvLSTM \cite{shi2015}, PredRNN \cite{predrnn}, PredRNN++ \cite{pmlr-v80-wang18b}, ConvGRU \cite{leinonen2021improvements}).

The state-of-the-art (SOTA) architectures for spatiotemporal weather forecasting \cite{leinonen2021improvements, sungbin_weather_stage2_2021, kwok_weather_stage2_2021} are memory intensive and require a substantially large number of parameters. In light of this, there is a need to develop light-weight, efficient, and improved accuracy spatiotemporal weather prediction models. 

In this research, we are motivated by the self-attention mechanism of transformers as it can capture the relationship between the input variables \cite{transformer_vaswani2017}. 
Indeed, vision transformers are gradually dominating computer vision applications, including dense prediction \cite{Cao2021_swin_unet,  ranftl2021vision_densePrediction}. These architectures  \cite{Liu2021_swin_transformer, Liu2021_video_swin_transformer, dong2021cswin, wang2021crossformer, Dosovitskiy2021_image_transformer} can be applied in weather prediction.
Solutions for this application involve an encoder (backbone, feature extractor) and a decoder (segmentor, predictor, forecaster) \cite{leinonen2021improvements, sungbin_weather_stage2_2021, kwok_weather_stage2_2021, shukla2013prediction}. This makes models designed for classification applicable as feature extractors. Also, decoders such as UNet \cite{unet}, FaPN \cite{huang2021fapn}, UperNet \cite{xiao2018unified}, SegFormer \cite{xie2021segformer} among others have been proposed for dense prediction (e.g., weather forecasting). Among all these decoders, SegFormer uses attention layers, while others are based on CNN.

\section{Methods}
\label{sec: methods}
\subsection{Overall Model Architecture}

\begin{figure*}[htbp!]
    \centering
    \includegraphics[width=\linewidth]{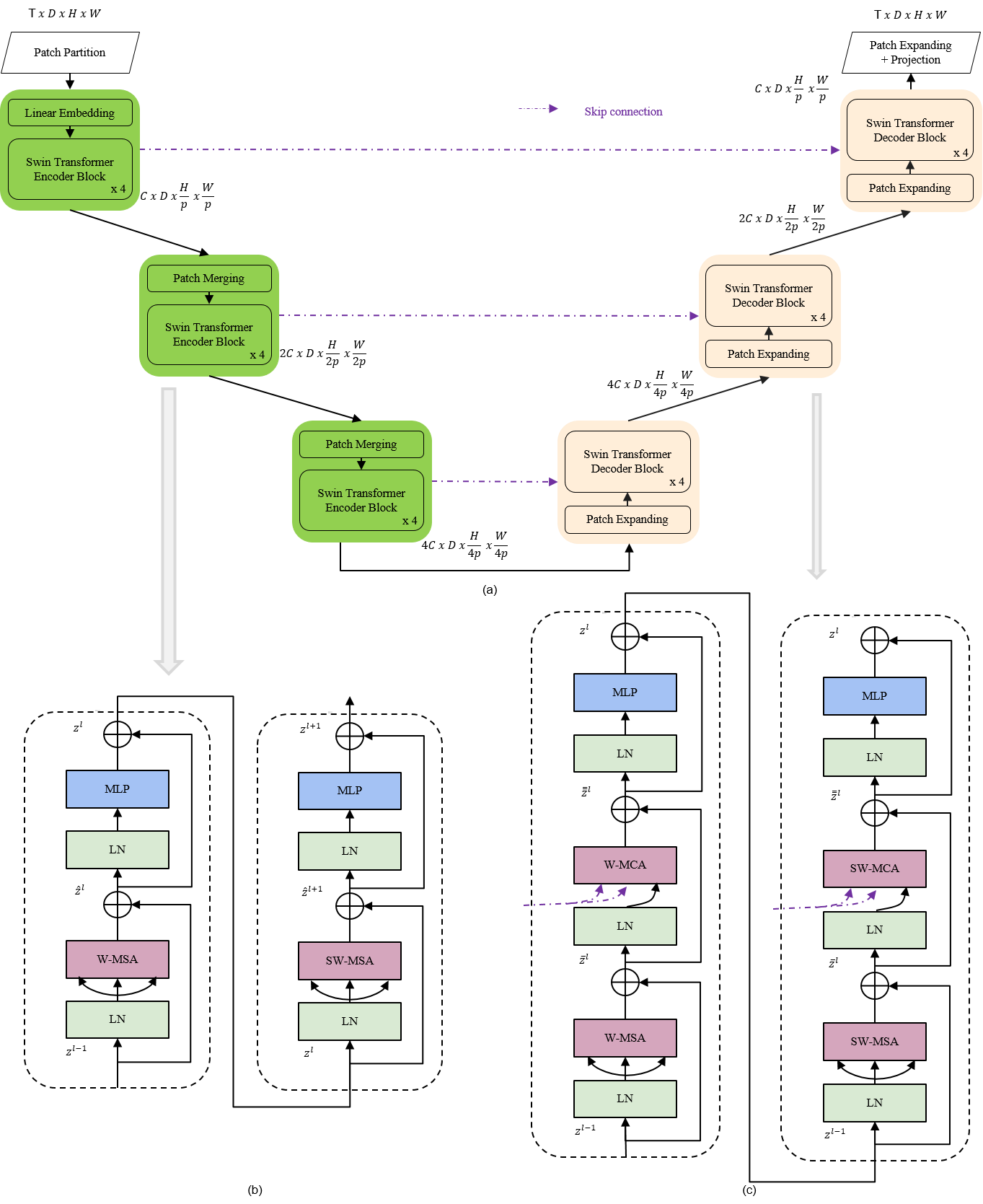}
    \caption{Spatiotemporal Encoder-Decoder architecture with cross attention in the decoder is shown in this diagram. (a) There are three transformer blocks and three decoder blocks in the model. Apart from the first encoder, which has linear embedding, each subsequent encoder block has a Patch Merging unit, followed by some Swin-Transformers. (b) The encoder's stacking of two concurrent self-attention blocks is shown, with shifted windowed attention always coming after non-shifted ones. (c) The decoder's cross-attention is demonstrated. Instead of using concatenation, which is usual in UNet topologies, the input from the skip connection and the decoding input are combined in a cross-attention layer. \cite{bojesomo2021spatiotemporal}}
    \label{fig: architecture}
\end{figure*}

The proposed model (see Fig. \ref{fig: architecture}) includes multiple stages for gradual downsampling of the spatial dimension in the encoder, which aid in capturing salient global representations. 
The encoder uses self-attention, and the decoder employs cross-attention to merge the skip connected input from the encoder with its main input  \cite{transformer_vaswani2017, bojesomo2021swindecoder}.
The developed model uses shifted window attention  \cite{Liu2021_video_swin_transformer}. This encourages the exploitation of inter-window relationships as local window attention is used to reduce the computational overhead (see Fig. \ref{fig: shifted window}).
Regular local transformer models exhibit a non-global correlation similar to convolutional networks. 
The inter-window relationship in shifted window transformer has the power to integrate the global correlation, even while being window-based. 

As shown in Fig \ref{fig: architecture}, the input enters the network through a patch embedding layer, while the final output of the model comes from a projected patch expanding layer.
The patch embedding layer converts the spatiotemporal inputs into tokens, while the final output is a projection of tokens to spatiotemporal format.
Three encoder and three decoder blocks are included in the model, each with four 3D transformer layers (encoder/decoder). 
The limited number of layers helps to avoid the need for enormous datasets to pre-train the transformer-based models \cite{Liu2021_swin_transformer, Liu2021_video_swin_transformer, Dosovitskiy2021_image_transformer, ranftl2021vision_densePrediction}.
The proposed model is composed of some layers and blocks; including the \textit{patch partitioning layer}, \textit{patch merging} layer, \textit{patch expanding} layer, and \textcolor{blue}{shifted window} transformer \textit{encoder} (as well as the \textit{decoder}) block.

\subsection{Patch Transformation Blocks}
\label{sec: patch transformation}
In the \textit{patch partitioning layer}, the spatiotemporal features are divided into patches and transformed using linear embedding. 
Two CNN layers are used to accomplish this. 
The first convolution has a stride value equal to its kernel size dimension to encourage patch partitioning, whereas the second convolution has a kernel size of 1 for linear embedding.

The \textit{patch merging layer} down-samples features using linear (fully-connected (FC) layer), making it a knowledge based operation and not traditional rule based ones (Average/Max pooling operation).
It typically flattens the features of each group of patches (2 x 2), followed by applying an FC-layer to convert the 4C-dimensional features to 2C-dimensional output, where C is the number of channels of the incoming features \cite{Liu2021_swin_transformer, Liu2021_video_swin_transformer}. 
On the decoding branch of the network, we have the \textit{patch expanding layer}, which works in  exactly the opposite way to the \textit{patch merging layer}, thus resulting in a learned up-sampling operation.


The projection head of the network includes a \textit{patch expanding layer} to recover the spatial dimensions, which may be updated in the network. 
This is followed by an FC-layer for channel dimension projection.


\subsection{Swin Transformer Block}
\label{sec: 3d swin tnx}
\begin{figure}[htbp!]
    \centering
    \includegraphics[width=1\columnwidth]{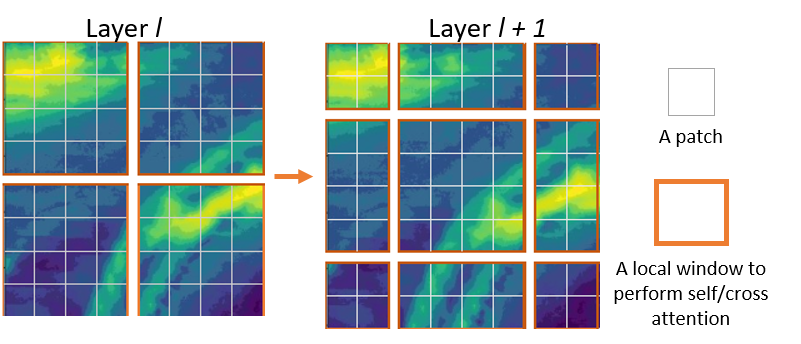}
    \caption{An outline of the local window grouping involved shifted window based architecture. The local window is shifted for a layer \textit{n + 1} following a regularly placed window in layer \textit{n}. 
    Since self(cross)-attention is window based, the shifting operation between layers provides the needed connection that is otherwise missing in local window-based attention operation. \cite{Liu2021_swin_transformer}}
    \label{fig: shifted window}
\end{figure}

A multi-head self-attention (MSA) \cite{Dosovitskiy2021_image_transformer} layer is followed by a feed-forward network (MLP) in the transformer layer used for computer vision tasks, with each of these layers preceded by layer normalization (LN) \cite{layer_norm}. 
Because of the spatiotemporal nature of the input, the 3D shifted window MSA is used in this study.
As highlighted in Figs. \ref{fig: architecture} (b) and (c), the Swin Transformer makes use of an interchange of sliding windows, where window (local) attention is next to another local but shifted window attention (See Fig. \ref{fig: shifted window}).
As an example, this arrangement leads to equation (\ref{eqn: attn}) for any two attention layers.

\begin{equation}
    \label{eqn: attn}
    \begin{split}
    \bar{z}^{l} = & \text{ W-MSA}(\text{LN}(z^{l-1})) + z^{l-1} \\
    z^{l} = & \text{ MLP}(\text{LN}(\bar{z}^{l})) + \bar{z}^{l} \\
    \\
    \bar{z}^{l+1} = & \text{ SW-MSA}(\text{LN}(z^{l})) + z^{l} \\
    z^{l+1} = & \text{ MLP}(\text{LN}(\bar{z}^{l+1})) + \bar{z}^{l+1}
    \end{split}
\end{equation}
where W-MSA and SW-MSA represent windowed multi-head self-attention and shifted window multi-head self-attention, respectively. 
In both W-MSA and SW-MSA, a relative position bias is used \cite{Liu2021_swin_transformer, Liu2021_video_swin_transformer}. 
However, the primary distinction between W-MSA and SW-MSA is a shift in window positioning before computing local attention within windowed blocks. 
In this research, we use (1, 7, 7) as the window size for all Swin transformers, with a shift size of 2 for the shift window versions.
Additionally, the MLP layers includes two FC-layers (eqn \ref{eqn: mlp}).

\begin{equation}
    \label{eqn: mlp}
    \text{MLP}(X) = f_1(f_2(X*W_1)*W_2)
\end{equation}
where $f_1$ and $f_2$ are the transfer functions of the two layers, respectively, $X \in \Re^{... \text{ x } d}$ is the input with $...$ being the other possible dimensions of the input, $W_1 \in \Re^{d \text{ x } 4d}$ is the weight matrix of the first (hidden) fully-connected layer and $W_2 \in \Re^{4d \text{ x } d}$ is the weight matrix of  (output) fully-connected layer.

\subsection{Shifted Window Cross Attention}
\label{sec: 3d swin cross tnx}

In the decoder, we use multi-head cross attention (MCA) which enables the interaction of the encoded tokens with the decoding ones. In contrast to MSA (Sec. \ref{sec: 3d swin tnx}) which uses the same input as the \textit{key}, \textit{query} and \textit{value}, MCA uses one input as the \textit{key} and \textit{value}, while using another as the \textit{query}. This ensures that we can explore the dependency of one input (\textit{query}) on the other parameters.

The use of multi-head cross attention only deals with interaction between the skip-connection (from the encoder) and the decoding input. There arises the need to further deal with self dependency in the form of self attention of the resulting output after MCA is applied before applying the feed-forward MLP network. The MSA layer is followed by another MCA layer in the decoder\cite{transformer_vaswani2017}. 
Such arrangement merges the skip-connected part with the decoding input. 
While this process is done using addition in LinkNet \cite{linknet_Chaurasia2017} and by concatenation in UNet \cite{unet}, we use multi-head cross attention \cite{transformer_vaswani2017}. The building blocks of our cross attention based decoding are preceded by layer normalization, an approach commonly used with Vision transformers \cite{Dosovitskiy2021_image_transformer, bojesomo2021spatiotemporal, bojesomo2021swindecoder, bojesomo2022swinunet3d}.

Similar to the encoder layers (Sec.\ref{sec: 3d swin tnx}), we use a 3D shifted window MCA in the decoder. Related to eqn. (\ref{eqn: attn}), the Swin tranformers in the decoder blocks alternately used self and cross attention, followed by shifted window self and cross attention as illustrated in equation (\ref{eqn: cross attn}).

\begin{equation}
    \label{eqn: cross attn}
    \begin{split}
    \bar{z}^{l} = & \text{ W-MSA}(\text{LN}(z^{l-1})) + z^{l-1} \\
    \bar{\bar{z}}^{l} = & \text{ W-MCA}(\text{LN}(\bar{z}^{l}), y) + \bar{z}^{l} \\
    z^{l} = & \text{ MLP}(\text{LN}(\bar{\bar{z}}^{l})) + \bar{\bar{z}}^{l} \\
    \\
    \bar{z}^{l+1} = & \text{ SW-MSA}(\text{LN}(z^{l})) + z^{l} \\
    \bar{\bar{z}}^{l+1} = & \text{ SW-MCA}(\text{LN}(\bar{z}^{l+1}), y) + \bar{z}^{l+1} \\
    z^{l+1} = & \text{ MLP}(\text{LN}(\bar{\bar{z}}^{l+1})) + \bar{\bar{z}}^{l+1} \\
    \\
    \bar{z}^{l+1} = & \text{ SW-MSA}(\text{LN}(z^{l})) + z^{l} \\
    z^{l+1} = & \text{ MLP}(\text{LN}(\bar{z}^{l+1})) + \bar{z}^{l+1}
    \end{split}
\end{equation}


\subsection{Swin Encoder}
The proposed model's encoder backbone network includes a multi-stage Video Swin transformer. We employ three stages, each with four 3D transformer blocks, followed by a patch merging layer.
\cite{Liu2021_swin_transformer, Liu2021_video_swin_transformer}. 
The multi-head self-attention (MSA) layer is used in the encoder, as described in section (\ref{sec: 3d swin tnx}) \cite{Liu2021_video_swin_transformer}.

\subsection{Swin Decoder}
We use a cross attention layer to encourage feature mixing before self-attention for subsequent interaction. The key parameter K and value parameter V in the cross-attention block is the skip connection from the encoder, where the continuing input from the patch expanding layer is the query parameter Q (See fig. \ref{fig: architecture}c).


\section{Experimental Results}
\label{sec: result}

\subsection{Data Description}
\label{sec: data}

\begin{figure*}[htbp!]
    \centering
    \includegraphics[width=\linewidth]{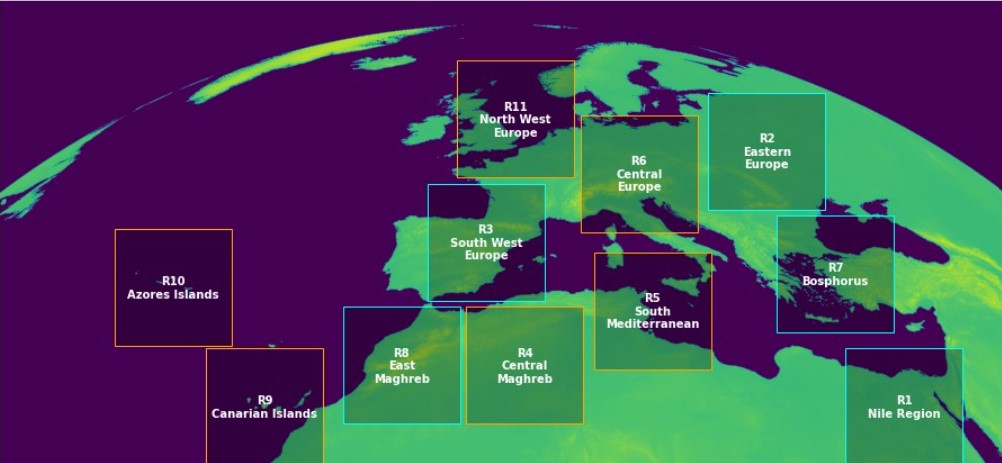}
    \caption{IEEE Big Data Weather4Cast 2021 Data Localization. The core challenge is shown in blue squares, while the regions in orange squares are for the transfer learning challenge\cite{ieee_bd_weather4cast2021}.}
    \label{fig: ieee-bd data}
\end{figure*}

The proposed system was tested on the challenging Traffic4cast 2021 \cite{cdceo_weather4cast, ieee_bd_weather4cast2021} weather dataset.
The dataset used was part of IEEE BigData Conference competition for weather movie snippet forecasting.
As shown in Fig. \ref{fig: ieee-bd data}, the dataset covers 11 regions including:
\begin{itemize}
    \item R1: Nile region (covering Cairo)
    \item R2: Eastern Europe (covering Moscow)
    \item R3: South West Europe (covering Madrid and Barcelona)
    \item R4: Central Maghreb (Timimoun)
    \item R5: South Mediterranean (covering Tripoli and Tunis)
    \item R6: Central Europe (covering Berlin)
    \item R7: Bosphorus (covering Istanbul)
    \item R8: East Maghreb (covering Marrakech)
    \item R9: Canary Islands
    \item R10: Azores Islands 
    \item R11: North West Europe (London, Paris, Brussels, Amsterdam)
\end{itemize}
This regions are grouped into two for both \textit{Core Challenge} and \textit{Transfer Challenge}. Regions used for the \textit{core challenge} are R1-R3, R7 and R8, and the data provided are divided into training, validation and test set. The \textit{transfer challenge} only has the test set for testing the transferability of the trained model on data from the remaining regions without prior training.

The data is presented in 256 x 256 weather image for various weather parameters as shown in Table \ref{tab: weather data}, categorized into five subgroups, with an area resolution of 4 km x 4 km per pixel. These weather images are captured 4 times per hour, amounting to one in 15 minute intervals.

\begin{table}[htbp!]
    \centering
    \caption{IEEE BigData Weather4Cast composition}
    \begin{threeparttable}
    \begin{tabular}{|p{0.25\columnwidth} | p{0.55\columnwidth}|}
        \hline
        Weather Product  & Weather variables \\
         \hline
         Temperature at Ground/Cloud (CTTH) & \textit{\textbf{temperature}, ctth\_tempe, ctth\_pres, ctth\_alti, ctth\_effectiv}, ctth\_method, ctth\_quality, ishai\_skt, ishai\_quality \\
         \hline
         Convective Rainfall Rate (CRR) & \textit{crr, \textbf{crr\_intensity}, crr\_accum}, crr\_quality \\
         \hline
         Probability of Occurrence of Tropopause Folding (ASII) & \textbf{asii\_turb\_trop\_prob}, asiitf\_quality \\
         \hline
         Cloud Mask (CMA) & \textit{cma\_cloudsnow, \textbf{cma}, cma\_dust, cma\_volcanic, cma\_smoke}, cma\_quality \\
         \hline
         Cloud Type (CT) & \textit{ct, ct\_cumuliform, ct\_multilayer}, ct\_quality \\
         \hline
    \end{tabular}
    \begin{tablenotes}
      \item Variables used in training our models are shown in \textit{italic} while target variables shown in \textbf{bold} are also used in training.
    \end{tablenotes}
    \end{threeparttable}
    \label{tab: weather data}
\end{table}

Some static context variables are also provided, including \textit{elevation} and \textit{longitude/latitude}. This static information is provided in a format of 256 x 256 pixels, similarly to the weather parameters. This information are unique to the specific regions and does not change over time.

The evaluation metric for these challenges (core and transfer challenges) considers the presence of missing values for each variable, and attempts to remove the dominance of any variable in the metric calculation. The scaling factor used to remove such (target) variable depending dominance is given by:
\begin{equation}
\label{eqn: persistence}
\small
\text {persistence}(v)=\left\{\begin{array}{ll}
0.03163512, & \text {if } v=\text { temperature } \\
0.00024158, & \text {if } v=\text { crr\_intensity } \\
0.00703378, & \text {if } v=\text { asii\_turb\_trop\_prob } \\
0.19160305, & \text {if } v=\text { cma }
\end{array}\right\}
\end{equation}
The evaluation metric using such persistence scaling leads to a value of 1 for persistence modeling and is given by:

\begin{equation}
\label{eqn: metric}
    \small
    Score_C = \frac{1}{D T R_C V} \sum_{d=1}^{D=36} \sum_{t=1}^{T=32} \sum_{r\in R_C} \sum_{v \in {V}} \frac{w(v)}{P_{r,v}} \sum_{p=1}^{P_{r,v}} (y_{v,p}^{r,d,t} - {\bar{y}}_{v,p}^{r,d,t}) 
\end{equation}
where $R_C$ represents the set of all regions involved for a given challenge, D is the total number of days in the testing set, T is the number of time steps, $w(v) := \frac{1}{persistence(v)}$ is the scaling factor attributed to each variable $v$ in the set of all target variables $V$. Also, $P_{r,v} = 256\times256 - N_r$ is used to account for missing data for a target variable $v$ in any region $r$, where $N_r$ is the number of missing data (i.e., empty pixels).

\subsection{Model Training}
Pytorch was used to implement the model shown in Figure \ref{fig: architecture}. In conjunction with the Adam optimizer  \cite{Kingma2015AdamAM}, we use the evaluation metric (eqn. \ref{eqn: metric}) as the loss function.
We trained the model using this loss function in a multitask setting, as the scaling factor involved ensures balance in the effects of each variable on the total loss.
The learning rate starts at 1e-4 and gradually reduced by half, when the validation set performance plateaued for more than 3 epochs.
The model was trained with a dedicated data augmentation scheme for dense prediction purposes. 
This augmentation pipeline includes random horizontal (\textit{RandomHorizontalFlip}) and vertical (\textit{RandomVerticalFlip}) flipping of the data, as well as randomly rotating the data block (90$^{\circ}$ \textit{RandomRotation})
The model training uses either the expected target data only or together with any combination of additional static and/or dynamic data available (Table \ref{tab: data categories}). The parameters of the models v2, v6, v7, v8 will be presented later in Table \ref{tab: model hyperparameters}. The additional data categories (i.e., static and dynamic) considered here are based on the fact that some weather variables are related, e.g., temperature and pressure. Also, the elevation map of a given location has some impact on weather fluctuations \cite{rangwala2012climate, kuhn2020elevation}.

\begin{table}[htbp!]
    \centering
    \caption{Data categories used in model training}
    \begin{threeparttable}
    \begin{tabular}{|p{0.15\columnwidth} | p{0.7\columnwidth}|}
        \hline
        Data type  & weather variables \\
         \hline
         target & temperature, crr\_intensity, asii\_turb\_trop\_prob, cma\\
         \hline
         static & elevation, latitude, longitude\\
         \hline
         dynamic & ctth\_tempe, ctth\_press, ctth\_effectiv, crr, crr\_accum, cma\_cloudsnow, cma\_dust, cma\_volcanic, cma\_smoke, ct, ct\_cumuliform, ct\_multilayer\\
         \hline
    \end{tabular}
    \end{threeparttable}
    \label{tab: data categories}
\end{table}

\subsection{Experiments}

We developed a forecasting model (Table \ref{tab: model hyperparameters}) that follows the architecture in Figure \ref{fig: architecture} with an embedding dimension of 48 and a patch-size of 4.
Training of the model configurations involved whether additional data will be used. One of the models was trained using only the target variables as listed in section \ref{sec: data}, while the remaining models include either other dynamic or static data, or a combination of the two together with the target variables (Table \ref{tab: model hyperparameters}). We used different combinations of data in model training to investigate the predictive capability of combining input data during training.

\begin{table}[htbp!]
    \centering
    \caption{Trained models performance comparison}
    \begin{threeparttable}
    \begin{tabular}{|c|c|c|c|ll|}
        \hline
        Model  & \multicolumn{2}{|c|}{Additional Data} & \#Parameters & \multicolumn{2}{|c|}{Performance} \\
         \hline
         version  & static & others & ~ & Core & Transfer \\
         \hline
         v2 &  No & No & 5.74M & 0.4936 & 0.4516\\ 
         v6 &  Yes & No & 5.78M & 0.4840 & 0.4516\\
         v7 &  No & Yes & 5.85M & 0.4810 & 0.4447\\
         v8 &  Yes & Yes & 5.87M & \textcolor{red}{0.4750} & \textcolor{brown}{0.4420}\\
        \hline
        \multicolumn{6}{|c|}{Top 3 models on the leaderboard} \\
        \hline
        ConvGRU \cite{leinonen2021improvements} (2021)& Yes & No  & 18M & \textcolor{blue}{0.4729} & \textcolor{blue}{0.4323}\\
        Dense UNet \cite{sungbin_weather_stage2_2021} (2021) & Yes &  Yes & 12M & \textcolor{brown}{0.4802} & \textcolor{red}{0.4376}\\
        Variational UNet \cite{kwok_weather_stage2_2021} (2021) & No &  Yes & 16M & 0.4857 & 0.4594\\
        \hline
    \end{tabular}
    \begin{tablenotes}
      \item \textcolor{blue}{blue}, \textcolor{red}{red}, and \textcolor{brown}{brown} colored results represents the 1$^{st}$, 2$^{nd}$ and 3$^{rd}$
    \end{tablenotes}
    \end{threeparttable}
    \label{tab: model hyperparameters}
\end{table}

As shown in Table (\ref{tab: model hyperparameters}), training with different combinations of inputs results in different model configurations in terms of the number of parameters (i.e., there is a change in the number of input channels).  


We compared our models with the best performing models for this dataset in Table \ref{tab: model hyperparameters} \cite{ieee_bd_weather4cast2021}. 
Our model has the least number of parameters and does not include ensembling, used by other models.

\subsection{Ablation Study}
We developed several forecasting models (Table \ref{tab: model variations}) that follow the architecture in Figure \ref{fig: architecture} with some changes in the hyperparameters (i.e., embedding dimension, patch size, weight decay, 90 degrees restricted rotation). As previously mentioned, model configuration training involves selecting whether additional data will be used. Some of the models were trained using only the target variables as listed in section \ref{sec: data}, while some models included either other dynamic or static data, or combinations of the two together with the target variables (Table \ref{tab: data categories} and \ref{tab: model variations}). 

\begin{table}[htbp]
    \centering
    \caption{Trained models configurations}
    \begin{threeparttable}
    \begin{tabular}{|c|c|c|c|c|c|}
        \hline
        model & \multicolumn{3}{|c|}{Hyperparameters} & \multicolumn{2}{|c|}{Additional Data}\\
         \hline
         version & embed-dim & patch-size & weight decay & static & dynamic \\
         \hline
         v0 & 16 & 2x2 & No & No & No \\
         v1 & 32 & 2x2 & No & No & No \\
         v2 & 48 & 2x2 & No & No & No \\
         \hline
         v3 & 48 & 2x2 & Yes & No & No \\
         v4 & 48 & 3x3 & Yes & No & No \\
         v5 & 48 & 4x4 & Yes & No & No \\
         v6 & 48 & 4x4 & Yes & Yes & No \\
         v7 & 48 & 4x4 & Yes & No & Yes \\
         v8 & 48 & 4x4 & Yes & Yes & Yes \\
        \hline
    \end{tabular}
    \begin{tablenotes}
      \small
      \item Starting with model v3, we included a weight decaying factor of 1e-6 during training to address possible overfitting.
    \end{tablenotes}
    \end{threeparttable}
    \label{tab: model variations}
\end{table}

\subsubsection{Embedding dimension variation}:
We trained several models that varied in the size of the embedding dimension, i.e., \{16, 32, 48\}. These dimensions were chosen to avoid an excessively large number of parameters as the intention was to develop parameter efficient models. All these models use a patch size of 2x2 without any additional data (static or dynamic) (Table \ref{tab: model variations}). The experimental results in Table \ref{tab: embed - dim} show that we can obtain an improvement in performance by increasing the embedding dimension.

\begin{table}[htbp]
  \caption{Model Performance with varying embedding dimension}
  \label{tab: embed - dim}
  \centering
      \begin{tabular}{|l|c|l|ll|}
        \hline
        Model & embed-dim & \#Parameters & Core & Transfer \\
        \hline
        v0 & 16 & 688,080 & 0.5015 & 0.4572 \\ 
        v1 & 32 & 2,574,688 & 0.4940 & 0.4530	\\ 
        v2 & 48 & 5,708,528 & \textbf{0.4936} & \textbf{0.4516}	\\ 
        \hline
      \end{tabular}
\end{table}

\subsubsection{Patch size variation}
We explored the effect of increasing the patch size used in the patch embedding layer of the model (Fig . \ref{fig: architecture}). In this experiment, we focused on the best performing model of Table \ref{tab: embed - dim}, which uses an embedding dimension of 48. The results of these experiments are shown in Table \ref{tab: patch size}, and indicate that a patch size of 4 achieved the best results. It is worth noting that increasing the patch size has a noteable improvement in terms of performance, while only incurring a very modest increase in the number of parameters.

\begin{table}[htbp]
  \caption{Model Performance with varying patch size}
  \label{tab: patch size}
  \centering
  \begin{threeparttable}
      \begin{tabular}{|l|c|l|ll|}
        \hline
        Model & patch size & \#Parameters & Core & Transfer \\
        \hline
        v3 & 2x2 & 5,708,528 & 0.5016 & 0.4516	\\ 
        v4 & 3x3 & 5,721,008 & 0.4974 & 0.4516	\\ 
        v5 & 4x4 & 5,738,480 & 0.4945 & 0.4516	\\ 
        \hline
      \end{tabular}
  \end{threeparttable}
\end{table}

\subsubsection{Using additional data}
It is worth noting that the earlier ablation experiments in this study solely used the target variables stated in Table \ref{tab: data categories}. In the current experiment, we trained the best performing model in Table \ref{tab: patch size}, i.e., with a patch size of 4x4 and an embedding dimension of 48, with different combinations of additional data. We also considered models with an embedding dimension of 32 for this experiment to reduce model size, and possibly enhance transfer-ability, and reduce overfitting. However, the experiments demonstrated that reducing the embedding dimension does not lead to improvements in model transfer, instead the model is self resilient to overfitting.

\begin{table}[htbp]
  \caption{Model Performance with additional data}
  \label{tab: model val results}
  \centering
  \begin{threeparttable}
      \begin{tabular}{|l|l|l|l|ll|}
        \hline
        Model & static & dynamic & \#Parameters & Core & Transfer \\
        \hline
        \multicolumn{6}{|c|}{Models with embedding dimension of 48} \\
        \hline
        v6 & Yes & No & 5,780,048 & 0.4840 & 0.4516	\\
        v7 & No & Yes & 5,847,632 & 0.4810 & 0.4516	\\
        v8 & Yes & Yes & 5,866,064 & \textbf{0.4750} & \textbf{0.4420}	\\
        \hline
        \multicolumn{6}{|c|}{Models with embedding dimension of 32} \\
        \hline
        v9 & Yes & No & 2,616,256 & 0.4845 & 0.4529	\\  
        v10 & No & Yes & 2,661,312 & \textit{0.4764} & 0.4481	\\  
        v11 & Yes & Yes & 2,673,600 & 0.4777 & \textit{0.4446}	\\ 
        \hline
      \end{tabular}
  \end{threeparttable}
\end{table}

\subsection{Qualitative Analysis}
\begin{figure*}[htbp!]
    \centering
    \includegraphics[width=1.0\linewidth]{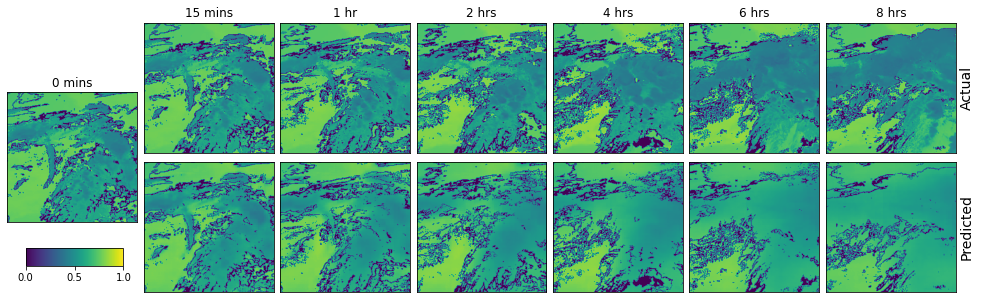}
    \caption{Comparing the \textit{temperature} forecast with groundtruth. The left side shows the situation before forecasting. On the right side, The first row shows the expected value in a sample of future times while the second row shows the predicted value. \textbf{Values normalized to the range (0, 1)}}
    \label{fig: temp}
\end{figure*}

\begin{figure*}[htbp!]
    \centering
    \includegraphics[width=1.0\linewidth]{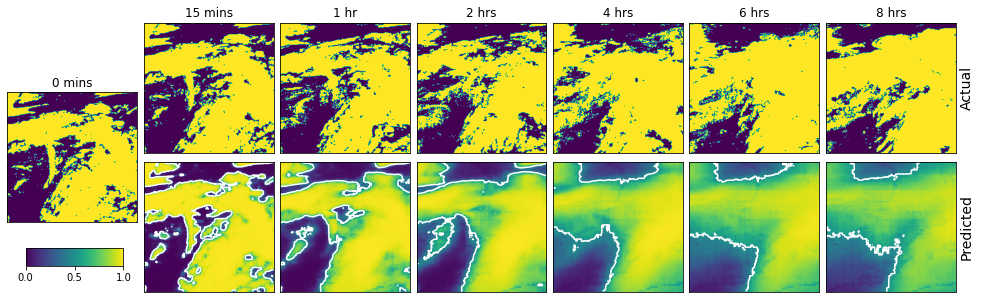}
    \caption{Comparing the \textit{cma} forecast with groundtruth. The left side shows the situation before forecasting. On the right side, The first row shows the expected value in a sample of future times while the second row shows the predicted value. \textbf{The contour line in white is used to show the binary threshold at 0.5.}}
    \label{fig: cma}
\end{figure*}

\begin{figure*}[htbp!]
    \centering
    \includegraphics[width=1.0\linewidth]{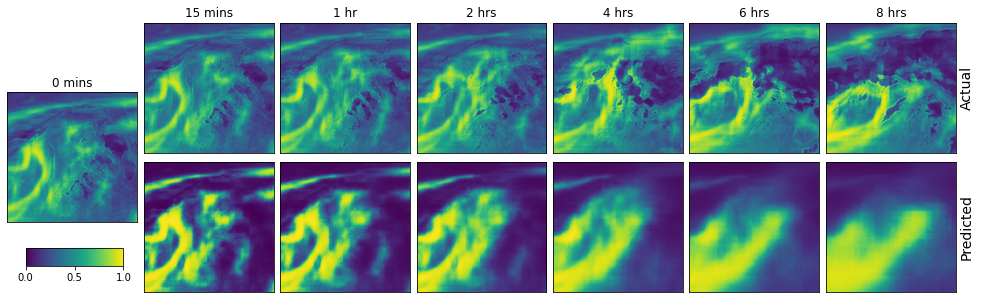}
    \caption{Comparing the \textit{asii\_turb\_trop\_prob} forecast with groundtruth. The left side shows the situation before forecasting. On the right side, The first row shows the expected value in a sample of future times while the second row shows the predicted value. \textbf{Values normalized to the range (0, 1)}}
    \label{fig: asii}
\end{figure*}

\begin{figure*}[htbp!]
    \centering
    \includegraphics[width=1.0\linewidth]{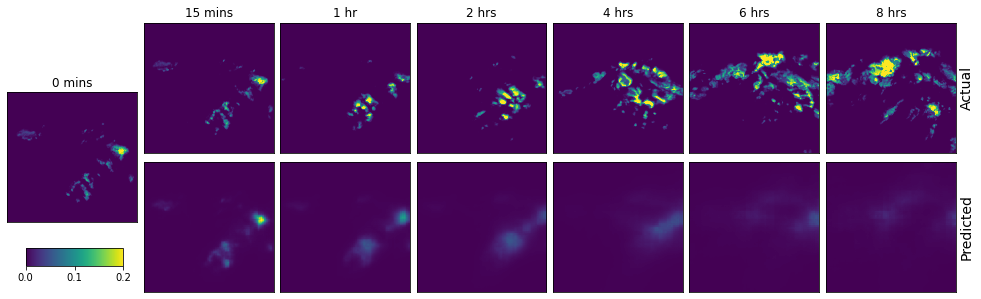}
    \caption{Comparing the \textit{crr\_intensity} forecast with groundtruth. The left side shows the situation before forecasting. On the right side, The first row shows the expected value in a sample of future times while the second row shows the predicted value. \textbf{Values normalized to the range (0, 1)}}
    \label{fig: crr}
\end{figure*}

We conducted a pictorial representation and analysis of the validation data based on the effect of time. Here, the variables are plotted individually to show the trend over the time of forecasting window. The weather conditions were compared to the ground truth as shown in Figs. \ref{fig: temp}, \ref{fig: asii}, \ref{fig: crr} and \ref{fig: cma}. 
This analysis shows that the model was able to accurately predict the weather variables quite well at start but its reliability decreases with time. 
As an example, the \textit{temperature} prediction in Fig. (\ref{fig: temp}) follows the expected value but slight blurriness towards the end to the prediction horizon.
This is similar to the observation in the analysis of \textit{cma} (fig. \ref{fig: cma}), \textit{asii\_turb\_trop\_prob} (\ref{fig: asii}), and \textit{crr\_intensity} (\ref{fig: crr}). The \textit{crr\_intensity} shown in figure (\ref{fig: crr}) indicate that the model can learn even in scarcity of non-zero values.



\section{Conclusions and Future Work}
\label{sec: conclusion}
We presented for the first time a short-time weather forecasting model that uses the 3D Swin-Transformer in a UNet architecture, which resulted in competitive results, i.e., a leaderboard score (scaled multitask MSE) of \textit{0.4750} and \textit{0.4420}, for the \textit{core} and \textit{transfer} challenges (IEEE Big Data Weather4Cast2021 \cite{ieee_bd_weather4cast2021}). 
The proposed model has only three blocks of Swin-transformers in both the encoder and decoder parts. It uses cross-attention in the decoder to merge data from the encoder with the upsampled decoding data. This ensures that the model focuses only on important information.
We intend to investigate different types of attention layers in the future. Similarly, we intend to investigate token mixing using hypercomplex networks such as sedenion \cite{bojesomo2020traffic}.

\section*{Acknowledgements}
  This work was supported by the ICT Fund, Telecommunications Regulatory Authority (TRA), Abu Dhabi, United Arab Emirates.

\bibliographystyle{IEEEtran}
\bibliography{references.bib, ml_weather.bib, JSTARS_2021.bib}

\end{document}